%% file: neurips_2025.tex
\documentclass{article}

 \usepackage[preprint]{neurips_2025}

\usepackage[utf8]{inputenc} % allow utf-8 input
\usepackage[T1]{fontenc}    % use 8-bit T1 fonts
\usepackage{hyperref}       % hyperlinks
\usepackage{url}            % simple URL typesetting
\usepackage{booktabs}       % professional-quality tables
\usepackage{amsfonts}       % blackboard math symbols
\usepackage{nicefrac}       % compact symbols for 1/2, etc.
\usepackage{microtype}      % microtypography
\usepackage{xcolor}         % colors
\usepackage[linesnumbered,ruled,vlined,boxed,noend]{algorithm2e}
\usepackage{amsmath}
\usepackage{amssymb}
\usepackage{multirow}
\usepackage{booktabs}
\usepackage{graphicx} % DO NOT CHANGE THIS
\usepackage{natbib}  % DO NOT CHANGE THIS AND DO NOT ADD ANY OPTIONS TO IT
\usepackage{caption} % DO NOT CHANGE THIS AND DO NOT ADD ANY OPTIONS TO IT

\title{Large–Small Model Collaborative Framework for Federated Continual Learning}

\author{%
  Hao Yu \\
  \small Southwestern University of Finance and Economics\\
  \texttt{yuhao2033@163.com} \\
  \And
  Xin Yang \thanks{Corresponding Author}\\
  \small Southwestern University of Finance and Economics\\
  \texttt{yangxin@swufe.edu.cn} \\
  \And
  Boyang Fan \\
  \small Southwestern University of Finance and Economics\\
  \texttt{fanby2014@126.com} \\
  \And
  Xuemei Cao \\
  \small Southwestern University of Finance and Economics\\
  \texttt{caoxuemei.qpz@gmail.com} \\
  \And
  Hanlin Gu\\
  \small WeBank\\
  \texttt{allengu@webank.com}
  \And
  Lixin Fan\\
  \small WeBank\\
  \texttt{lixinfan@webank.com}\\
  \And
  Qiang Yang\\
  \small Hong Kong Polytechnic University\\
  \texttt{profqiang.yang@polyu.edu.hk}\\
}

\begin{document}

\maketitle

\begin{abstract}
% AAAI creates proceedings, working notes, and technical reports directly from electronic source furnished by the authors. To ensure that all papers in the publication have a uniform appearance, authors must adhere to the following instructions. AAAI creates proceedings, working notes, and technical reports directly from electronic source furnished by the authors. To ensure that all papers in the publication have a uniform appearance, authors must adhere to the following instructions. 
% AAAI creates proceedings, working notes, and technical reports directly from electronic source furnished by the authors. To ensure that all papers in the publication have a uniform appearance, authors must adhere to the following instructions. AAAI creates proceedings, working notes, and technical reports directly from electronic source furnished by the authors. To ensure that all papers in the publication have a uniform appearance, authors must adhere to the following instructions. AAAI creates proceedings, working notes, and technical reports directly from electronic source furnished by the authors. To ensure that all papers in the publication have a uniform appearance, authors must adhere to the following instructions. 

% Federated Foundation Models (FMs) exhibit superior performance in Federated Continual Learning, where each participant needs to continually learn on a private sequence of tasks while mitigating Spatial-Temporal Catastrophic Forgetting. 

Continual learning (CL) for Foundation Models (FMs) is an essential yet underexplored challenge, especially in Federated Continual Learning (FCL), where each client learns from a private, evolving task stream under strict data and communication constraints.
Despite their powerful generalization abilities, FMs often exhibit suboptimal performance on local downstream tasks, as they are unable to utilize private local data.
Furthermore, enabling FMs to learn new tasks without forgetting prior knowledge is inherently a challenging problem, primarily due to their immense parameter count and high model complexity. 
In contrast, small models can be trained locally under resource-constrained conditions and benefit from more mature CL techniques. 
To bridge the gap between small models and FMs, we propose the first collaborative framework in FCL, where lightweight local models act as a dynamic bridge, continually adapting to new tasks while enhancing the utility of the large model. Two novel components are also included: \textbf{Small Model Continual Fine-tuning} is for preventing small models from temporal forgetting; \textbf{One-by-One Distillation} performs personalized fusion of heterogeneous local knowledge on the server.
Experimental results demonstrate its superior performance, even when clients utilize heterogeneous small models.

\end{abstract}


\input{sects/intro}

\input{sects/related_work}

\input{sects/problem}

\input{sects/methodology}
\input{sects/experiments}

\input{sects/conclusion}

\bibliographystyle{plainnat}
\bibliography{aaai2026}

\end{document}

%% file: sects/intro.tex
\section{Introduction}

% 第一段介绍FM的能力很强，并且FMs的出现
Foundation Models (FMs), also commonly referred to as large models, have emerged as a groundbreaking force in the field of artificial intelligence \cite{zhou2024comprehensive}. Prominent FMs like GPT-4 and LLaMa, owing to their large parameter sizes, possess outstanding capacity to address diverse tasks across fields such as natural language processing (NLP) and computer vision (CV) \cite{awais2025foundation}. Due to their strong generalization capabilities, applying them to Federated Learning (FL) can help address many challenges such as Non.I.I.D. and data heterogeneity \cite{ren2025advances}.

\begin{figure}[t]
    \centering
    \includegraphics[width=0.82\linewidth]{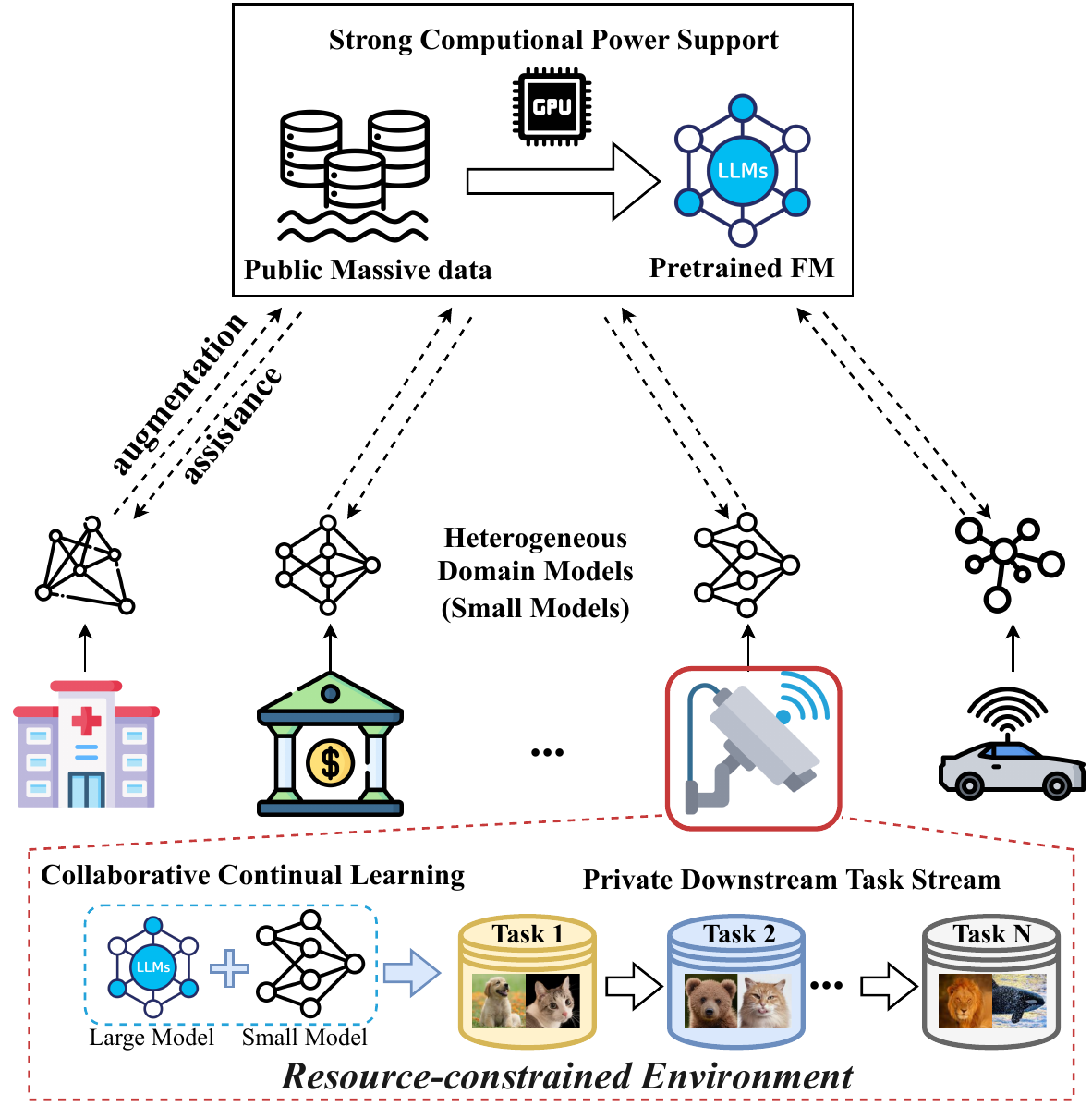}
    \caption{A large FM, pre-trained on massive public data, is deployed to the client side to assist small model in CL over private downstream task streams. In return, the small models enhance the FM with their private knowledge through FL. }
    \label{Intro}
\end{figure}

In real-world settings, data and tasks evolve over time, static FL is no longer sufficient to meet the demands of changing environments. Federated Continual Learning (FCL) is a practical paradigm which allows each participant in FL to continually learning on a private sequence of tasks without Spatial-Temporal Catastrophic Forgetting (STCF). STCF refers to the phenomenon where clients not only gradually forget the knowledge they have previously learned during the process of Continual Learning, but also forget their local knowledge after federated aggregation \cite{yang2024federated}. And FCL is well-suited for resource-constrained scenarios like Edge-AI, as it enables incremental integration of knowledge from different clients over time without retraining the model, thus reducing computation and communication overhead \cite{yang2024efficient, zhang2022cross}.

% 介绍FMs在FL下遇到的问题。1. FM无法利用Local 数据的知识  2.客户端资源受限，无法训练整个FM  3.在客户端本地进行FCL更是难上加难，大模型本身就难以CL，在资源受限的联邦环境下进行FCL更是困难
However, the deployment of FMs in FCL settings introduces several fundamental challenges. First, due to privacy constraints, client data is confined to local devices and cannot be directly shared with FMs, preventing these models from leveraging valuable knowledge from private domains \cite{kang2023grounding}. Moreover, clients are often resource-constrained, lacking the computational capacity to train entire large models. In most cases, they can only utilize these models without any form of adaptation or personalization, which leads to suboptimal performance on local downstream tasks \cite{pan2024federated}. Finally, when clients are exposed to a continual stream of evolving tasks, the difficulties intensify: not only is CL itself a non-trivial problem for large models \cite{ostapenko2022continual}, but ensuring that FMs can incrementally acquire new knowledge without forgetting previously knowledge becomes especially challenging under resource-constrained conditions. Therefore, how to enable CL in Federated Foundation Models has become one of the most challenging problems \cite{fan2025ten}.

To overcome these challenges, a straightforward solution is to integrate `powerful but heavy' FMs with `lightweight and flexible' Domain Models (DMs) into a Large–Small Model Collaborative Framework for FCL, as illustrated in Fig. \ref{Intro}.  These DMs, being trained entirely on local data, are thus better able to leverage privacy-sensitive knowledge. Consequently, by harnessing the global representation power of FMs and the localized, fine-grained feature extraction capabilities of DMs, the framework is well-equipped to more effectively satisfy the demands of client-specific tasks.

However, it also brings new challenges. \textbf{First and foremost}, a key challenge lies in how to effectively bridge the gap between DMs and FMs, such that FM's performance on downstream tasks can be enhanced through the DM.
\textbf{Second}, while DMs are compatible with mature CL techniques such as EWC \cite{kirkpatrick2017overcoming} and LwF \cite{li2017learning}, which makes them more practical for deployment in resource-constrained federated environments, it remains unclear how to transfer this CL capability from DM to FM, so that FM can benefit without being directly involved in the CL process itself. \textbf{Third}, under this collaborative framework, another important question arises: what form of client-side knowledge should be uploaded to the server in order to improve generalization through FL, while preserving privacy and efficiency?

To this end, we propose the first collaborative large-small model framework in FCL, termed Fed-LSCL, where lightweight local models perform CL with the assistance of a large FM. In turn, these local models continually fine-tune the FM without requiring direct CL or fine-tuning of the FM itself. Fed-LSCL comprises three novel components. \textbf{Large-Small Model Collaborative Training} allows the small model to act as a dynamic bridge, continually adapting to new tasks while enhancing the large model's overall utility.
\textbf{Small Model Continual Fine-tuning} enables DMs to generate consistent parameters for fine-tuning FMs on past samples.
On the server side, \textbf{One-by-One Distillation} sequentially distills each client’s parameter generator using the generators of the other clients as teacher models, deliberately avoiding fusion of the heterogeneous local CNNs.
% On the server side, \textbf{One-By-One Distillation} is performed solely on each client's parameter generator without aggregating the local CNNs, thereby avoiding fusing heterogeneous knowledge. 
This not only reduces communication overhead but also enhances privacy by avoiding transmission of local model parameters. Main contributions are summarized as follows:

% #####################################################

\begin{itemize}
    \item A Large–Small Model Collaborative Framework is first designed for Federated Continual Learning. It effectively combines the global generalization ability of FM with the fine-grained local representational capacity of DMs, enabling more effective local task learning.
    \item We adopt a small model, which is more easy to CL, to incrementally fine-tune a pre-trained ViT, thereby avoiding direct continually fine-tuning on the large and complex FM. Based on that, we then design a novel CL strategy for the small model and federated aggregation strategies tailored to this framework, effectively mitigating the effects of spatial-temporal catastrophic forgetting.
    \item Extensive experiments not only demonstrate the state-of-the-art performance of Fed-LSCL, but also highlight its superior computational efficiency and privacy preservation compared to conventional fine-tuning approaches.
\end{itemize}

%% file: sects/related_work.tex
\section{Related Work}

\subsection{Pretrained Model-based FCL}
Traditional FCL methods, such as FedWeiT \cite{yoon2021federated} and GLFC \cite{dong2022federated}, employ small models to achieve greater flexibility in FCL tasks. However, the limited capacity of these models also renders them more susceptible to the effects of spatial-temporal catastrophic forgetting, thereby limiting their generalization capabilities.

In contrast, methods based on FMs utilize frozen large pre-trained models to achieve more generalized spatial-temporal knowledge fusion \cite{yu2025handling}. PiLoRA \cite{guo2024pilora} proposes a parameter-efficient LoRA variant designed to reduce communication overhead. 
LoRM \cite{salamiclosed} explores how low-rank matrix factorization can effectively fine-tune large models in FCL. FedMGP takes into account the multi-granularity expression of knowledge, promoting the spatial-temporal integration \cite{yu2024personalized}.

These methods demonstrate excellent performance in retaining knowledge from both new and old tasks. However, their high computational overhead makes them difficult to deploy in resource-constrained environments. Furthermore, they exhibit limited flexibility when adapting to continuously arriving data streams, as most rely on a query function to select and load relevant parameters.

\subsection{Large–Small Model Collaborative Learning}
To fully leverage the generalization capabilities of large models and the flexibility of small models, current research is actively exploring collaborative strategies between them.

% Small Model-Enhanced Large Models
TinyLLM \cite{tian2025beyond} leverages knowledge distillation from a large model to empower a lightweight small model with near-large model performance. Another approach is the Weak-to-Strong paradigm \cite{burns2024weak}, which utilizes small models as ``weak supervisors" to progressively guide large models in learning more complex tasks. HuggingGPT \cite{xie2023doremi} enables a large language model to select and invoke different expert models (including small models) based on user instructions to complete a series of complex tasks. \cite{ye2022zerogen} empowers small models with zero-shot or few-shot generalization to new tasks, leveraging large models even with insufficient data. However, these methods typically do not involve joint training of small and large models, nor do they account for more realistic continual fine-tuning scenarios. 

To achieve collaborative learning in FCL, this work focuses on combining the excellent representational capabilities of large models with the lightweight and flexible characteristics of small models. It utilizes local small models to act as a dynamic bridge, continually adapting to new tasks while enhancing the utility of the large model.

%% file: sects/problem.tex
\section{Problem Definition}

Given a FL setting with $k$ heterogeneous clients, denoted as $\mathcal{A}=\{A_1, A_2, \ldots, A_k\}$, and a central server $S$. Each client $A_i$'s local dataset $\mathcal{D}_i$ comprises two distinct sets of classes: private classes $\mathcal{C}_v^i$ and public classes $\mathcal{C}_p$. Specifically, $\mathcal{C}_v^i$ represents classes whose data samples are exclusively present on client $A_i$, ensuring data privacy and locality. The public classes $\mathcal{C}_p$ are shared across all clients, where we enforce a strict non-overlapping constraint among the data instances of $\mathcal{C}_p$ to maintain data uniqueness. 

To characterize extreme spatial heterogeneity, the number of the public class can be set to zero, i.e., $|\mathcal{C}_p|=0$. Experimental validation for this extreme case is conducted on ImageNet-R.
To characterize temporal heterogeneity, the task sequence of client $A_i$ is denoted by $\mathcal{T}_i = \{T_i^1, T_i^2, \ldots, T_i^{N}\}$, where $N$ is the total number of tasks assigned to $A_i$. Each task contains an equal number of classes, but the class sets are mutually exclusive across tasks.

At current task $n$, the local small model $\theta_i^{n-1}$ of client $A_i$ will continual learn on the current task $T_i^n$ without forgetting previous knowledge from $T_i^1$ to $T_i^{n-1}$. The goal of training $\theta_i^{n}$ is not only to perform continual adaptation to new tasks but also to enhance the fixed large foundation model $\Theta$ to get better performance on the local downstream tasks. After local training, the server will collect and aggregate these local models to generate the global small model $\theta_g^{n}$ with more generalized global knowledge. Therefore, the overall objective is formulated as:
\begin{equation}
\begin{split}\small
    \min_{\theta_g^{n}}F\Big(\underbrace{\ell_{1}(\Theta(\theta_g^{n}),\{T_1^t\}_{t = 1}^{n-1}), \cdots, \ell_{k}(\Theta(\theta_g^{n}),\{T_k^t\}_{t = 1}^{n-1})}_{\text{old task loss}}, \\ \underbrace{\ell_{1}(\Theta(\theta_g^{n}),T_1^{n}),\ell_{k}(\Theta(\theta_g^{n}),T_k^{n})}_{\text{new task loss}}\Big),
    \end{split}
\end{equation}
where $\ell_{k}(\Theta(\theta_g^{n}),\{T_k^t\}_{t = 1}^{n-1})$ and $\ell_{k}(\Theta(\theta_g^{n}),T_k^{n})$ are the loss functions of client $k$ corresponding to the old tasks and the new task, respectively. $F$ is the aggregation mechanism. $\Theta(\theta_g^{r})$  represents the foundation model $\Theta$ after undergoing fine-tuning guided by $\theta_g^{n}$.

%% file: sects/methodology.tex
\section{Methodology}

\subsection{Overall Framework}
To address the spatial-temporal catastrophic forgetting and bridge the gap between DMs and FMs, we design a collaborative learning mechanism, called Fed-LSCL, where the lightweight small model $\theta_i$ not only performs continual adaptation to new tasks but also learns to generate sample-aware \textit{adapter parameters} for the fixed $\Theta$ through the parameter generator $\phi_i^r$. Unlike traditional fine-tuning approaches that directly update large models and incur high computation and storage costs, we freeze the parameters of $\Theta$ and instead optimize the small model to produce sample-specific adapter parameters that are injected into $\Theta$ to adjust its representations. This collaborative optimization framework allows the \textit{small model to act as a dynamic bridge}, continually adapting to new tasks while enhancing the utility of the large frozen model. And the server can only aggregate small model $\theta_i$ to achieve spatial knowledge fusion, avoiding the need to transmit or update the large model $\Theta$, which significantly reduces communication and computation costs.

\begin{figure*}[t!]
    \centering
    \includegraphics[width=0.9\linewidth]{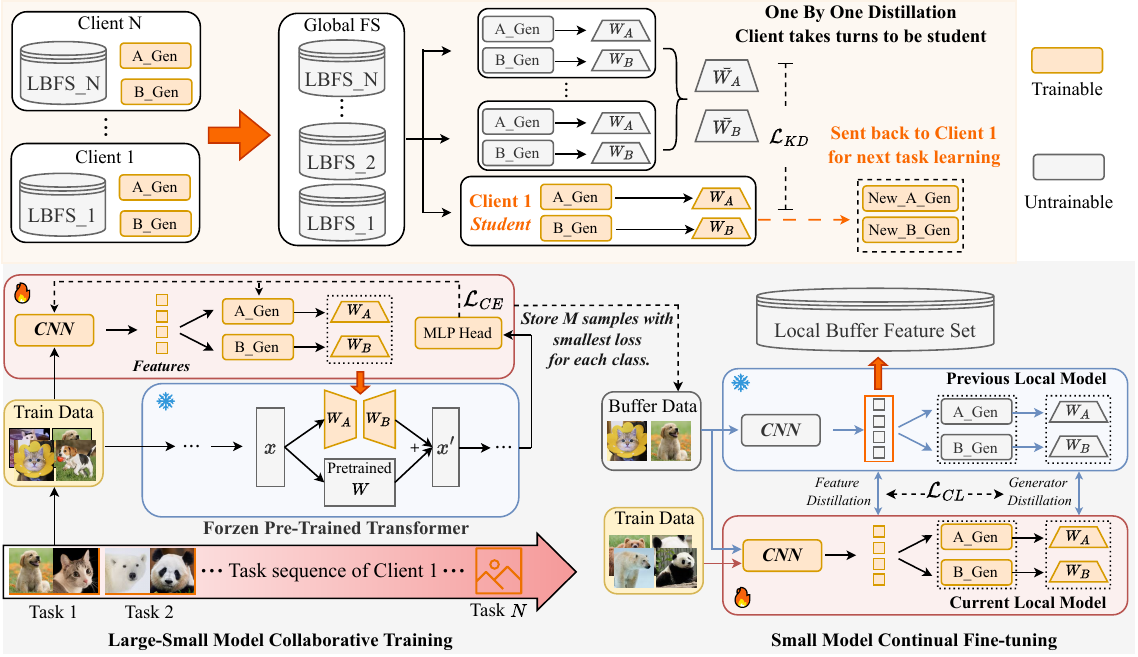}
    \caption{The proposed Fed-LSCL framework consists of three components: \textbf{Large-Small Model Collaborative Training} transforms continual fine-tuning of the large model into parameter generation via a small model, reducing client-side overhead; \textbf{Small Model Continual Fine-tuning} preserves the small model’s ability to generate past parameters, addressing temporal forgetting; \textbf{One-by-One Distillation} fuses diverse parameter generators across heterogeneous clients to mitigate spatial forgetting.}
    \label{method}
\end{figure*}

The overall framework of Fed-LSCL is shown in Fig. \ref{method}. and the algorithm is summarized in Alg. \ref{alo1}. 

\begin{algorithm}[t!]
\caption{Fed-LSCL Algorithm.}
\label{alo1}
  % \SetAlgoLined
  \KwIn{$k$ clients $\mathcal{A} = \{A_i\}_{i=1}^k$ with their own task sequence $\mathcal{T}_i = \{T_i^n\}_{n=1}^N$, a pre-trained frozen ViT $\Theta$ without classification head, local classification head $H_i$.}
  
  \KwOut{Personalized fused $\{\bar{\phi_i^{n}}\}_{i=1}^k$ for each client.}
  % \BlankLine
  
  Initialization\;
  \While{task number n $\leq$ N}{
        \For{each client $A_i, 1 \leq i \leq k$}{
            \rm{\textbf{Stage 1: Collaborative Training:}}
            
            \For{each $\{x, y\} \in T_i^n$}{
                $E$ $\gets$ $\theta_i^{n}(x)$\;
                $W_{i}$  $\gets$ $\phi_i^{n}(E)$ \;
                $\Theta_i^n$ $\gets$ \rm{Fine-Tuning}($W_{i}$, $\Theta$) \;
                $\mathcal{L}_{ce}$ $\gets$ $CrossEntropy(H_i(\Theta_i^n(x)),y)$ \;
                \rm{Optimize}($\mathcal{L}_{ce},{H}_i,\phi_i^{n},\theta_i^{n}$)\;
            }

            \rm{\textbf{Stage 2: Continual Fine-tuning:}}
            
            \For{each $\{\hat{x}, \hat{y}\} \in $ LocalBuffer $B_i^{n-1}$}{
                 $\{\hat{E},\hat{E'}\}$ $\gets$ $\{\theta_i^{n}(\hat{x}),\theta_i^{n-1}(\hat{x})\}$\;
                 $\{\hat{W_{i}},\hat{W'_{i}}\}$  $\gets$ $\{\phi_i^{n}(\hat{E}),\phi_i^{n-1}(\hat{E'})\}$ \;
                 $\mathcal{L}_{cl}$ $\gets$ $\mathcal{L}_{kd}(\hat{E},\hat{E'}) + \mathcal{L}_{kd}(\hat{W_{i}},\hat{W'_{i}})$ \;
                 $\hat{\Theta_i^n}$ $\gets$ \rm{Fine-Tuning}($\hat{W_{i}}$, $\Theta$) \;
                 $\mathcal{L}_{ce}$ $\gets$ $CrossEntropy(H_i(\hat{\Theta_i^n}(x)),\hat{y})$ \;
                 \rm{Optimize}($\mathcal{L}_{ce},\mathcal{L}_{cl},{H}_i,\phi_i^{n},\theta_i^{n}$)\;
            }

            $B_i^{n}$ $\gets$ \rm{UpdateBuffer}($T_i^n,B_i^{n-1}$) \;
            $FS_i^n$  $\gets$ \rm{ConstructFeatureSet}($\theta_i^{n},B_i^{n}$) \;
            UploadToServer( $\phi_i^{n},FS_i^n$)
 
        }

        \rm{\textbf{Stage 3: Server aggregation:}}

        \For{each client $A_i, 1 \leq i \leq k$}{
            $\bar{\phi_i^{n}}$ $\gets$ \rm{OBOD}($\phi_i^{n},\{\phi_j^{n}\}_{j \neq i}^k,\{FS_j^n\}_{j \neq i}^k$) \;
            \rm{Sent $\bar{\phi_i^{n}}$ to $A_i$ for next task training.}
        
        }
                
    }
\end{algorithm}

\subsection{Large-Small Model Collaborative Training}
% Given the extreme difficulty of directly fine-tuning pre-trained foundation models in resource-constrained settings, the prospect of enabling them to engage in continual learning is even more challenging. Some CL methods based on pre-trained models, such as L2P and DualPrompt, achieve continual learning of large models by loading different parameters through a Query-Function.

Enabling pre-trained FMs to acquire continual learning capabilities remains highly challenging. On one hand, their massive parameter sizes make it impractical to perform full model training on resource-constrained environments such as federated clients. On the other hand, existing methods for equipping large models with continual learning abilities are still in their infancy. Most approaches rely on specific query functions to select and activate discrete parameter subsets for the model, such as in L2P \cite{wang2022learning}, DualPrompt \cite{wang2022dualprompt}, and similar techniques.

Motivated by this, we naturally propose leveraging small models to continually adapt and modulate pre-trained large models—not only because of their low computational overhead, but also due to the maturity of CL techniques based on small models. We further observe that such a collaborative framework is particularly well-suited for FCL. On one hand, small models can be trained locally on privacy-sensitive data, preserving user privacy; on the other hand, compared to conventional FL approaches, it offers stronger privacy protection by avoiding full model sharing.

In this paper, we assign each client with the same pre-trained ViT $\Theta$ as a foundation model, which remains \textit{\textbf{fixed and untrainable}} throughout the entire FCL process. This large model $\Theta$ serves as a powerful, general-purpose feature extractor and a robust backbone for classification. And each client possesses a CNN as its own small model, denoted as $\theta_i$. Note that these CNNs can have different architectures, such as ResNet-18, VGG-16, and so on. 

For a given input sample and its label $(x,y) \in T_i^n$, it first passes through the local small model's feature extractor $\theta_i^{n}$ to obtain an intermediate representation $E$.
\begin{equation}
    E = \theta_i^{n}(x).
\end{equation}
This representation $E$ is then fed into the parameter generator $\phi_i^{n}$, which generates a set of sample-specific parameters, denoted as $W_i$. In this paper, $W_i$ serves as the LoRA matrix \cite{hulora} to fine-tune the pre-trained ViT $\Theta$, thereby enhancing the classification performance.

Let $\Theta_i^n$ denote the ViT which is fine-tuned by $W_i$ and let $H_i$ represent the local classification head. To jointly train the small model parameters $\theta_i^{n}$, the adapter generator $\phi_i^{n}$, and the local classification head $H_i$ while keeping the ViT $\Theta$ fixed, we define the loss for client $A_i$ on task $T_i^n$ as:

\begin{equation}
\mathcal{L}_{i,\text{total}}^n =\mathbb{E}_{(x,y) \sim T_i^n} \left[ \mathcal{L}_{\text{CE}}\left( H \left( \Theta_i^n \left(\phi_i^{n}(E) \right) \right), y \right) \right].
\end{equation}

This collaborative optimization framework allows the \textit{small model to act as a dynamic bridge}, continually adapting to new tasks while enhancing the utility of the large model. 

% And the server can only aggregate small model $\theta_i$ to achieve spatial knowledge fusion, avoiding the need to transmit or update the large model $\Theta$, which significantly reduces communication and computation costs.

\subsection{Small Model Continual Fine-tuning}

\subsubsection{Continual Fine-tuning}
Thanks to the proposed collaborative framework, we can avoid the challenging and resource-intensive task of directly performing continual learning on the complex large model itself. Instead, we only need to focus on designing corresponding methods for the smaller model and its associated parameter generator, thereby enabling the effective continual fine-tuning of the large model.

Drawing inspiration from the classical method LwF \cite{li2017learning}, we save the local CNN from the previous time step and its corresponding parameter generator on the local client, denoted as $\theta_i^{n-1}$ and $\phi_i^{n-1}$ respectively. For every data sample 
$(\hat{x}, \hat{y})$ in the buffer $B_i^{n-1}$, it is processed by both the previous local model and the current local model. Specifically, the previous CNN processes it to yield features $\hat{E}'$ and generate matrix $\hat{W}_i'$, while the current CNN processes it to produce features $\hat{E}$ and generate matrix $\hat{W}_i$. And the continual loss can be summarized as:
\begin{equation}\small
\begin{split}
    \mathcal{L}_{\text{CL}} = \frac{1}{|B_i^{n-1}|} \sum_{(\hat{x}, \hat{y})} \left[\lambda_E  \left\| \hat{E} - \hat{E}' \right\|_2^2 +\lambda_W \left\| \hat{W}_i - \hat{W}_i' \right\|_2^2\right],
\end{split}
\end{equation}
where $\lambda_E$ and $\lambda_W$ are the hyper-parameter. Utilizing this loss, we aim to ensure that the local small model extracts consistent features for past samples and that the local parameter generator produces similar matrices. This mechanism empowers the local small model with continual ability, thereby enabling it to continually fine-tune the large model.

\subsubsection{Sample Selection}
For important sample selection, we adopt a class-balanced strategy to maintain the local data buffer in this work. Specifically, after completing the large-small model collaborative training for the current task $T_i^n$, we select the top-$M$ samples $(\hat{x}, \hat{y})$ with the lowest classification loss $\mathcal{L}_{\mathrm{CE}}\left( \Theta_i^n(\phi_i^n(x)), y \right)$ within each class $c$ from the current task’s training data, and store them to update the local buffer $B_i^n$. This selection strategy ensures class balance and prioritizes stable, representative samples.

Note that these two sub-components \textit{can both be replaced by existing continual learning methods for small models}, as long as they can enable the parameter generator to generate similar matrices for fine-tuning the large model.

\subsection{One By One Distillation}
Unlike most existing traditional FL methods, Fed-LSCL only requires the upload of local feature set extracted by the local CNN from buffer data and the local parameter generator $\phi_i^n(x)$ to achieve knowledge fusion across different clients. However, merely generating a single global model often leads to degraded performance of the global model on local tasks, as some local-specific parameters are altered during the global aggregation process. Therefore, we propose \textit{One By One Distillation}, which generates a unique, personalized global model for each client on the server, thereby effectively preventing performance degradation.

Denote the local buffer feature set of client $A_i$ at task $n$ as $FS_i^n$. All participating clients $A_k$ upload their respective local buffer feature sets $FS_k^n$ along with their corresponding parameter generators $\phi_k^n$ to the server. For the distillation process on the server, the parameter generator $\phi_k^n$ of each client $A_i$ is designated as the student model. $\phi_j^n$ from all other clients $(j\neq i)$ serve as teacher models. Crucially, when teacher $\phi_j^n$ starts to the distillation, the knowledge transfer is performed by utilizing features derived from $FS_j^n$. This process aims to distill the collective knowledge from diverse teachers into $\phi_i^n$, while simultaneously constraining $\phi_i^n$ to prevent significant deviation. Therefore, the server-side distillation loss for client $A_i$ is:
\begin{equation}\small
\label{eq:server_distillation_loss_multi_teacher}
\bar{\phi}_i^n = \arg\min_{\phi} \sum_{j \neq i} \mathbb{E}_{x \in FS_j^n} \left[
\left\| \phi(x)  -\phi_j^n(x) \right\|_2^2
\right] + \lambda \left\| \phi - \phi_i^n \right\|_2^2.
\end{equation}
The first term is the distillation loss, while the second term constrains parameter updates from excessively deviating from the original parameters, thereby preserving local task performance. Here, $\lambda$ is a hyperparameter.

\subsection{Equivalence Analysis of One-by-One Distillation}

We now provide a theoretical interpretation of the proposed \textit{One-by-One Distillation} (O2D) method. Recall that in O2D, each client $A_i$ serves as a \textit{student}, and all other clients $A_j$ ($j \neq i$) serve as \textit{teachers}. The student model $\phi_i$ is optimized to distill knowledge from the outputs of $\phi_j$ using buffered features $x \in FS_j^n$.

% For a given feature $x$, let $z_j = \phi_j(x)$ be the output (e.g., adapter weights or logits) of teacher $j$, and let $z_i = \phi_i(x)$ be the student output. The O2D loss minimizes the KL divergence between $z_i$ and $z_j$, averaged across teachers:

% \begin{equation}
% \mathcal{L}_{\text{O2D}}(\phi_i) = \sum_{j \ne i} \mathbb{E}_{x \sim BF_j^n} \left[ 
% \mathcal{D}_{\text{KL}} \left( z_j \parallel z_i \right)
% \right] + \lambda \|\phi_i - \phi_i^0\|_2^2,
% \end{equation}

% where $\phi_i^0$ is the original local generator before server-side distillation. For simplicity, we consider the case where $z_j, z_i \in \mathbb{R}^d$ are softmax probabilities or normalized logits.

Let $\mathcal{Z}_i = \{ \phi_j^n(x) \}_{j \ne i}$ denote the set of teacher outputs for a given $x$. Then, minimizing $\sum_j \|z_i - z_j\|^2$ is equivalent to projecting $z_i$ onto the convex hull of $\mathcal{Z}_i$:

\begin{equation}
\min_{z_i} \sum_{j \ne i} \| z_i - z_j \|_2^2
\quad \Longleftrightarrow \quad 
z_i^\star = \arg\min_{z \in \text{Conv}(\mathcal{Z}_i)} \|z - \bar{z}\|_2^2,
\end{equation}
where $\bar{z} = \frac{1}{k-1} \sum_{j \ne i} z_j$ is the mean teacher output. Thus, the O2D objective implicitly encourages the student output $z_i$ to approach a \textbf{convex combination of teacher outputs}, i.e., a knowledge consensus, while the regularization term $\|\phi - \phi_i^n\|_2^2.$ ensures that the updated generator remains close to the original local model, preserving personalization.

\textbf{Conclusion.} The One-by-One Distillation process can be viewed as performing a convex projection of the student output onto the subspace spanned by teacher models, thereby achieving a personalized yet generalized update in the space of parameter generators.

%% file: sects/experiments.tex
\section{Experiments}
\subsection{Experimental Setup}
\subsubsection{Datasets}
To effectively capture the impact of spatial-temporal data heterogeneity, we partition the data as follows: 

For \textbf{ImageNet-R} \cite{hendrycks2021many}, we exacerbate spatial data heterogeneity by assigning 40 unique private classes to each client, with no public classes. Similarly, each task for ImageNet-R also consists of 8 classes. For \textbf{CIFAR-100} \cite{krizhevsky2009learning}, each client is assigned 15 private classes, which are exclusive to itself. This setup results in 25 public classes that are shared across clients. Consequently, each client handles data for a total of 40 classes (15 private + 25 public), aligning with the 5 tasks, each containing 8 classes. Note that we employ a Dirichlet distribution for assigning data within the public classes to create a non-IID data distribution among clients.

\subsubsection{Baseline Methods} We compare Fed-LSCL with following method: FedAvg \cite{mcmahan2017communication}, FedProx \cite{li2020federated}, FedViT \cite{dosovitskiy2020image}, FedL2P \cite{wang2022learning}, FedDualP \cite{wang2022dualprompt}, GLFC \cite{dong2022federated}, TARGET \cite{zhang2023target}, MFCL \cite{babakniya2024data}, FedMGP \cite{yu2024personalized}, PILORA \cite{guo2024pilora} and LoRM \cite{salami2025closedform}. Brief introduction of baseline methods is provided in Appdenix A.

\subsubsection{Implementation Details}
All experiments are conducted with three random seeds (42, 1999, 2024), and the averaged outcomes are reported. For all evaluated methods, the number of clients is fixed at five, and the interval between task increments is consistently set to five communication rounds. Optimization is performed using the Adam optimizer with a learning rate of $10^{-3}$. The entire training process is conducted sequentially on a single NVIDIA GPU RTX-3090. Local CNNs are Resnet-18 \cite{he2016deep}.

% \newpage
\begin{table*}[!t]
\setlength{\tabcolsep}{3pt} 
\centering
\tiny
\begin{tabular}{c|c|ccccc|ccccc}
\toprule
\multirow{2}{*}{Algorithm}  & \multirow{2}{*}{Type} & \multicolumn{5}{c}{CIFAR-100 Task ID} & \multicolumn{5}{c}{Imagenet-R Task ID} \\

 & &  1& 2 & 3 & 4 & 5 & 1& 2 & 3 & 4 & 5 \\

\midrule
FedAvg  &  AISTATS 2017 & 43.925 &50.678 & 57.369 & 55.519 & 61.230  & 37.778 &35.427 & 35.536 &35.811 &36.751  \\

% FedAvg+EWC &  &0.86 & 1.00 & 1.00 & 1.00 & 1.00 \\
FedProx  & MLSys 2020 &23.703 & 22.803 & 26.146 & 22.152 & 23.664 & 20.238 & 19.762 & 19.730 & 18.984 & 17.882 \\

FedAvg+ViT &  ICLR 2020 & 70.237 & 70.139 & 71.413 & 66.027 & 67.389 & 68.284 & 59.818 & 57.396 & 59.867 & 57.950 \\ 

FedAvg+L2P & CVPR 2022 &28.427 & 29.962 & 29.304 & 25.440 &25.703 & 27.183 & 27.602 & 24.839 & 25.103 & 26.518\\ 

FedAvg+DualP & ECCV 2022 &31.772 & 42.839 & 52.839 & 39.005 & 46.398 & 23.530 & 26.630 & 26.402 & 30.237 & 32.038 \\

\midrule
GLFC & CVPR 2022 & 82.055 & 63.129 & 73.437 & 64.230 & 64.867 &	61.934 & 67.134 & 67.235 & 71.713 & 57.293 
 \\

TARGET & ICCV 2023 & 54.023  & 41.491 & 32.289 & 13.913 & 15.902 
 & 39.993 & 15.038  & 16.028 &  17.506 & 16.103 \\ 

MFCL & NeurIPS 2024 & 46.723 & 16.236 & 10.603 & 14.616 & 13.539& 28.856 &14.576&16.265 & 13.347 & 13.833 \\

FedMGP &  KDD 2024& 90.263 & 85.307 &90.736 & 89.223 & 82.236 & 77.339 & 76.830 &78.092 & 75.612  & 75.407 \\

PILoRA & ECCV 2024 & 84.961 & 78.874 & 72.578 & 71.104 & 64.684 & 77.107 & 68.595 &	64.974 &	61.348 & 56.723\\

LoRM  & ICLR 2025 & 93.773 &	88.684 &	85.160 &	84.556 &	80.534 &87.823 & 80.802 & 79.618 &	75.480 & 71.918\\

\midrule

Ours (Fed-LSCL) & --- &  \textbf{97.735} &	\textbf{97.419} &	\textbf{96.597} &	\textbf{96.544} &	\textbf{96.932} 
&  \textbf{91.178}	& \textbf{89.155}	 & \textbf{86.985}&\textbf{87.382} &\textbf{88.150}\\

% Ours-w/o OBOD &  & \\

% Ours-w/o SMFT &  &  \\

% Ours-w/o LSCL & &  \\

\bottomrule
\end{tabular}
\caption{Task-wise accuracy of the global model on local test sets with 5 class-incremental tasks (each task with 8 classes).}
\label{tab:cifar100}
\end{table*}

\subsection{Experimental Results}
\subsubsection{Main Results}
We use the accuracy of the global model on the current local test set as the metric in Tab. \ref{tab:cifar100}.

Given the strong spatial-temporal data heterogeneity in the data distribution, FedAvg, FedProx, and most CL combined FL methods exhibit suboptimal performance. Conversely, FedAvg+ViT demonstrates acceptable performance, which we attribute to its strategy of aggregating only the classification head while keeping backbone frozen.

For traditional FCL methods that do not leverage pre-trained models, GLFC exhibited excellent performance on CIFAR-100 but performed poorly on complex image datasets such as ImageNet-R. Concerning MFCL and Target, both rely on data generation to overcome catastrophic forgetting. Given that all images in our experiments are resized to $3\times224\times224$, training high-quality data generators at this size poses a significant challenge. Consequently, the performance of these two methods fell short of expectations.

For the FCL methods that leverage pre-trained models,  these approaches generally show strong performance. Specifically, FedMGP, which utilizes multi-granularity prompts, and LoRM, which employs low-rank matrix to fine-tune ViT, both demonstrate excellent results across two datasets. They achieve nearly 90\% accuracy on CIFAR-100 and approximately 80\% accuracy on ImageNet-R. In contrast, PILoRA's performance gradually declines as the number of tasks increases, with its performance on the final task being comparable to that of GLFC.

Fed-LSCL demonstrates the best performance on both datasets, showing a significant improvement particularly on the ImageNet-R dataset. Furthermore, its performance does not decline noticeably as the number of tasks increases, which fully highlights our method's effectiveness.

\subsubsection{Ablation Studies}
\setlength{\tabcolsep}{3pt} 
\begin{table}[]
    
    \centering
    \begin{tabular}{ccc|cc}
    \toprule
         L-S Colla. & SM-CF & O2D & CIFAR-100 & Imagenet-R  \\
         \midrule
         $\checkmark$ & $\checkmark$ &$\times$ & 89.396	 & 80.780 \\
         $\checkmark$ & $\times$ & $\checkmark$ & 50.298 & 46.648 \\
         $\checkmark$ & $\times$ & $\times$ & 49.235	 & 45.377 \\
         $\times$ & $\times$ & $\times$ & 29.873	& 27.298\\
    \bottomrule
    \end{tabular}
    \caption{Ablation study on three key components.}
    \label{tab:Ablation}
\end{table}
To evaluate the contribution of each component, ablation studies are conducted on the three key components proposed in this paper. The results in Tab. \ref{tab:Ablation} are the average performance of the final global model when back-tested on all local tasks. \textit{L-S Colla.} signifies Large-small collaboration, \textit{SM-CF} refers to small-model continual fine-tuning, and \textit{O2D} denotes one-by-one distillation.

The results in Table \ref{tab:Ablation} highlight the importance of all three components of our framework. In particular, Large-Small Model Collaborative Training and Small Model Continual Fine-tuning play a critical role in overcoming spatial-temporal catastrophic forgetting in FCL.

\subsubsection{Sensitivity Studies}
We conduct a sensitivity analysis on which ViT block is fine-tuned by the lightweight model, as well as on the number of buffered samples stored per class. The metric in both figures is \textit{the accuracy of the aggregated global model at different time points on all local tasks}.

\begin{figure}[htbp!]
    \centering
    \includegraphics[width=0.8\linewidth]{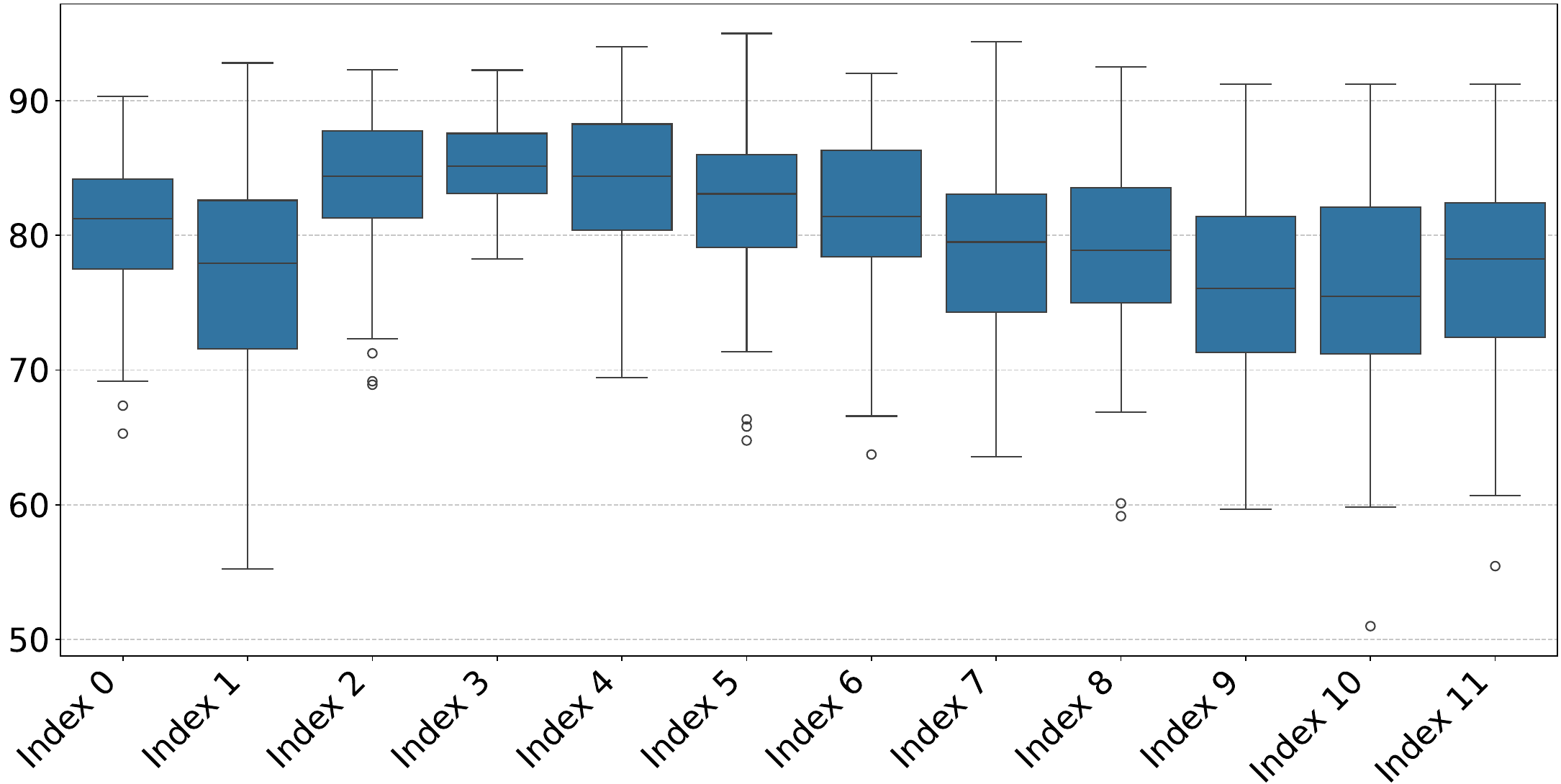}
    \caption{Sensitivity analysis of small model fine-tuning within different blocks on accuracy (ImageNet-R).}
    \label{fig:index}
\end{figure}

\begin{figure*}[htbp!]
    \centering
    \includegraphics[width=1\linewidth]{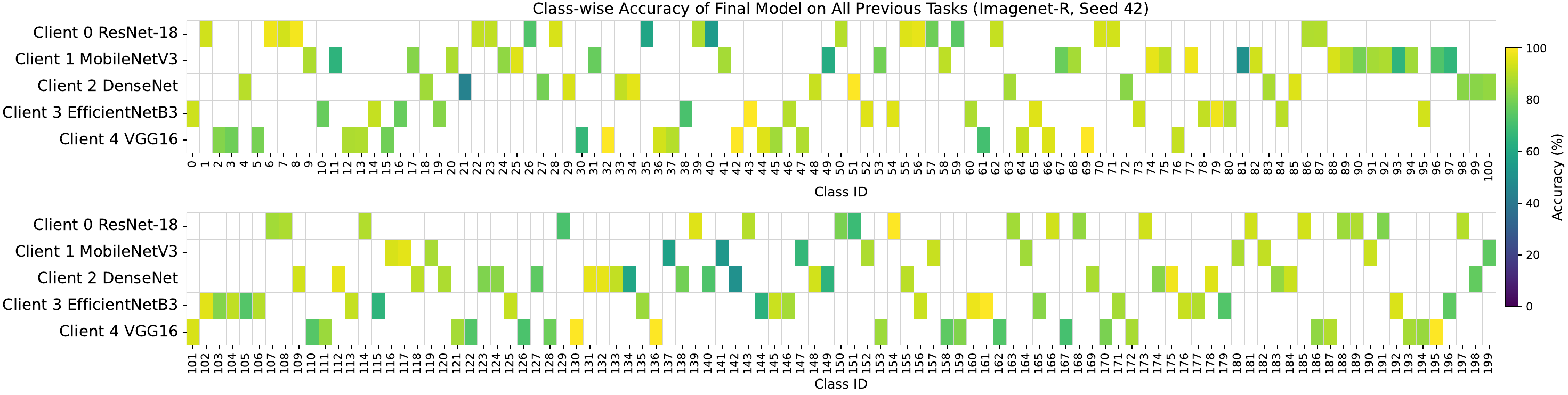}
    \caption{Client-wise accuracy of final model on all previous tasks.}
    \label{fig:class-wise}
\end{figure*}

As shown in Fig. \ref{fig:index}, when the small model is fine-tuned on the shallow blocks of ViT (e.g., Blocks 0 to 5), the global model demonstrates robust performance across all cumulatively learned tasks. Notably, at Block 3, the model's performance appears relatively stable and consistent. However, when the small model is fine-tuned on deeper ViT blocks (such as Blocks 9 to 11), a significant degradation in model performance is observed, and the performance curves also display greater variability.

\begin{figure}[h]
    \centering
    \includegraphics[width=0.8\linewidth]{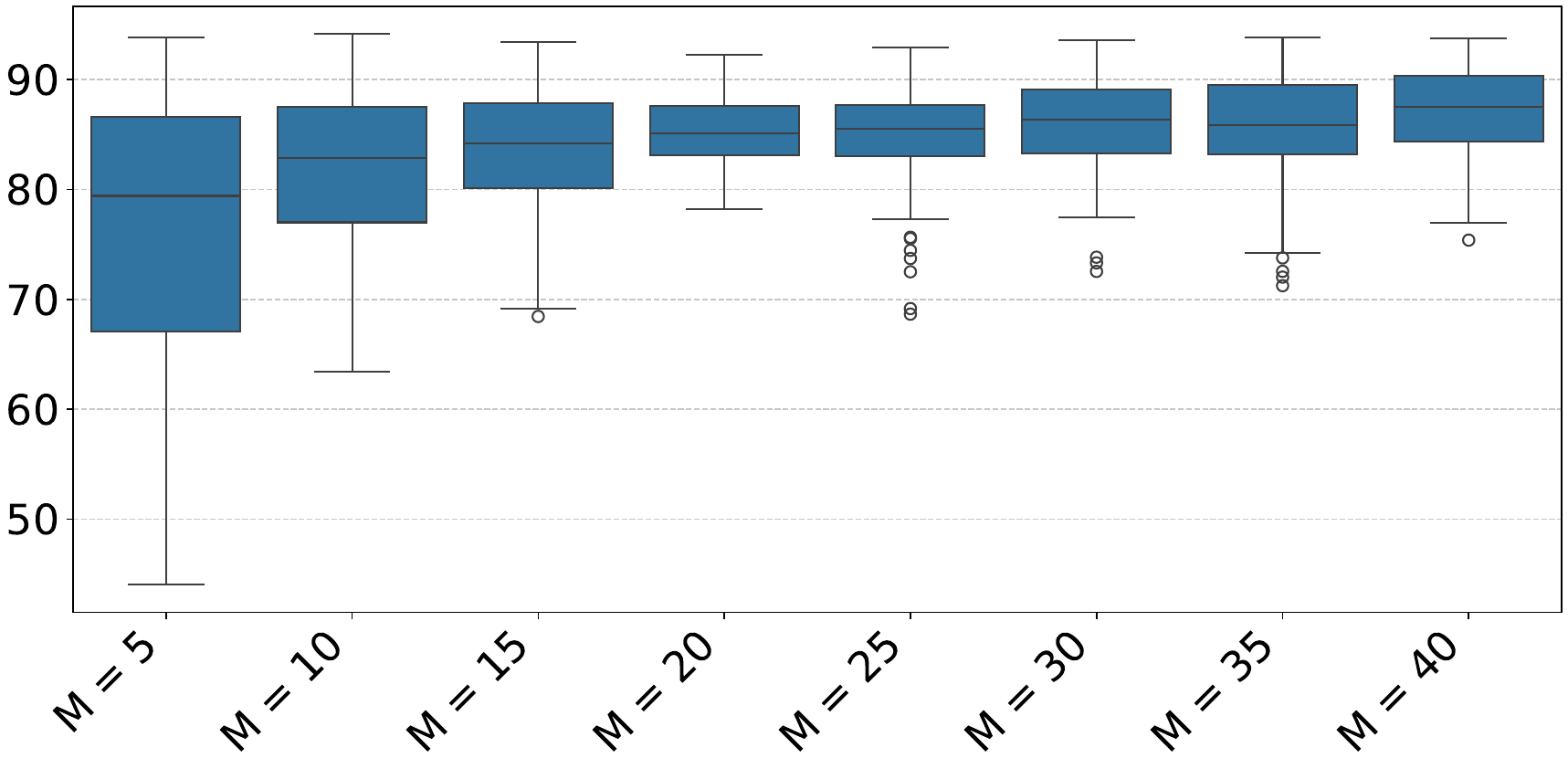}
    \caption{Sensitivity analysis of buffer size per class (ImageNet-R).}
    \label{fig:num}
\end{figure}

Fig. \ref{fig:num} illustrates Fed-LSCL's performance across different buffer sizes per class. The results indicate that Fed-LSCL achieves satisfactory performance by buffering merely 20 representative samples per class, thus underscoring the efficiency in local memory consumption.

\subsubsection{Heterogeneity of Local CNN}
To demonstrate the robustness of Fed-LSCL under heterogeneous local model settings, we assign a different CNN architecture to each client. Specifically, client 0 uses ResNet-18 \cite{he2016deep}, client 1 uses MobileNetV3 \cite{howard2019searching}, client 2 uses DenseNet \cite{huang2017densely}, client 3 uses EfficientNet-B3 \cite{tan2019efficientnet}, and client 4 uses VGG-16 \cite{simonyan2015very}. Fig. \ref{fig:class-wise} presents the class-wise accuracy of the final global model on all local tasks. The results demonstrate that Fed-LSCL, despite the heterogeneity in local models and computational resources across clients, consistently achieves strong performance. This highlights the effectiveness of the proposed large-small model collaboration strategy in FCL settings.

\subsection{Privacy \& Efficiency Analysis}
\subsubsection{Privacy Protection} 
In Fed-LSCL, local client $A_i$ only upload the parameter generator $\phi_i^n$ and the local buffer feature set $FS_i^n$ of the buffer data, while the locally trained CNN is not uploaded. Unlike traditional FL, Fed-LSCL keeps the locally trained CNN on the client device. Specifically, the local buffer feature set is a distilled version of the client's local data, processed by the client's local CNN. Since the server cannot access the local CNN's network parameters or architecture, it's significantly more difficult to infer private data from these feature sets. Similarly, the parameter generator is only responsible for converting these features into partial parameters for fine-tuning the large model. Therefore, Fed-LSCL inherently offers strong privacy protection.

\subsubsection{Communication \& Computational Cost}
In Fed-LSCL, the primary communication overhead consists of only two parts: the local buffer feature set and the parameter generator. The parameter generator itself is a linear layer designed to convert features extracted by the local CNN into parameters suitable for fine-tuning the large model. As for the local buffer feature set, its size is determined by the feature dimension, the number of data classes, and the count of samples cached per class. As for local computational overhead, it only involves the number of parameters in the local CNN and the parameter generator. This makes it highly efficient.

%% file: sects/conclusion.tex
\section{Conclusion}
This paper introduces Fed-LSCL, a novel small-large model collaborative learning framework designed for Federated Continual Learning (FCL). This approach effectively addresses critical FCL challenges, including spatial-temporal catastrophic forgetting, resource constraints, and privacy protection. It allows a small model to act as a dynamic bridge, continually adapting to new tasks while enhancing the utility of a large, frozen pre-trained model. By combining the global representation power of FMs with the localized, fine-grained feature extraction capabilities of DMs, this framework is able to more effectively meet the requirements of client-specific tasks. This design significantly reduces communication overhead and enhances privacy by preventing the transmission of private local model parameters. Experimental results showcase its state-of-the-art performance across datasets, particularly noting a significant improvement on ImageNet-R. The method also exhibits robust performance even when clients utilize local small models with diverse architectures, showcasing its resilience to model, data, and computational heterogeneity.